# Development of the user-friendly decision aid Rule-based Evaluation and Support Tool (REST) for optimizing the resources of an information extraction task

Guillaume Bazin (1,2), Xavier Tannier (1), Fanny Adda (3), Ariel Cohen (4), Akram Redjdal (1,2), Emmanuelle Kempf (1,3)

1 Sorbonne Université, Inserm, Université Sorbonne Paris Nord, Laboratoire d'Informatique Médicale et d'Ingénierie des Connaissances pour la e-Santé, Limics, Paris, Fr
2 ESIEE Paris Engineering school, Noisy-le-Grand, Fr
3 Assistance Publique – Hôpitaux de Paris, Henri Mondor Teaching Hospital, Department of Medical oncology, Créteil, Fr
4 Assistance Publique – Hôpitaux de Paris, IT Department, Paris, Fr

**Corresponding author**

Emmanuelle Kempf
Department of Medical oncology
Henri Mondor Teaching Hospital
Assistance Publique – Hôpitaux de Mondor
1 rue Gustave Eiffel
94010 Créteil, France
Emmanuelle.kempf@aphp.fr
P : +33 1 49 81 45 31

- Author contributions

    All authors made a significant contribution to the concept, design, acquisition, analysis or interpretation of data; drafted the article or revised it critically for important intellectual content; approved the final version of the article for publication; agreed to be accountable for all aspects of the work and resolved any issues related to its accuracy or integrity

- Statements and Declarations
    - Declaration of conflicting interest

        The authors have no conflicts of interest to disclose.

    - Funding statement

        This project had no funding.

    - Ethical approval and informed consent statements

        Ethical approval was not required.

    - Data availability statement

        The YouTube video tutorial (https://youtu.be/v58IqJxVnCo) summarize the specifications of REST

**Abstract**
Rules could be an information extraction (IE) default option, compared to ML and LLMs in terms of sustainability, transferability, interpretability, and development burden. We suggest a sustainable and combined use of rules and ML as an IE method. Our approach starts with an exhaustive expert manual highlighting in a single working session of a representative subset of the data corpus. We developed and validated the feasibility and the performance metrics of the REST decision tool to help the annotator choose between rules as a by default option and ML for each entity of an IE task. REST makes the annotator visualize the characteristics of each entity formalization in the free texts and the expected rule development feasibility and IE performance metrics. ML is considered as a backup IE option and manual annotation for training is therefore minimized. The external validity of REST on a 12-entity use case showed good reproducibility.

## 1. Introduction

In recent years, artificial neural network algorithms have become the reference for information extraction (IE) tasks, due to the unprecedented leap in computing power and the massive increase in the amount of data collected available for training (1). Despite the current wave of large language model (LLM) development in IE tasks, such techniques might still lack effectiveness (2–4), while rule-based IE models are old-fashioned systems which keep being used currently, despite a potential lack of interest from the scientific community (5). Indeed, a literature review showed that 66% of IE studies relied on rule systems solely (6). Rule-based systems are useful options in specialized fields, such as the processing of biomedical texts (7). For example, rules are likely to be hybridized with machine learning (ML) IE techniques to perform automatic pre-annotation with quite satisfactory performance metrics, particularly when the terminology in question is restricted and formally homogeneous (8,9). In specific use cases, rule-based systems could be associated with IE performance metrics equal or superior to ML and LLM techniques (10–12). Under specific text conditions, rules seem to be a widely used and relevant alternative IE technique to ML and LLM (13).

Yet, despite a decade-long availability of both rule-based and ML techniques, developers may lack the tools to determine which of these systems is the optimal one for a dedicated IE task. In this article, we hypothesized that IE tasks could be optimized thanks to a combination of both rules and ML techniques, at the level of each entity of interest. Based on a lower carbon footprint and annotation workload, and on a better interpretability and transferability, we consider rules as a default IE method (14). We suggested a checklist of criteria to help the annotator to choose between both options according to each entity's formal text characteristics. We developed and evaluated the feasibility and the reproducibility of the REST user friendly visualization and decision aid tool on a 12-entity use case.

The contributions of this paper are:
- To define qualitative and quantitative textual criteria, as a checklist, for each entity of interest, assessing the feasibility of rules for an IE task;
- To develop a user-friendly visualization and decision aid tool to determine if rules could be a reliable IE method;
- To perform an external validation of such method on REST.

## 2. Conceptual framework and methodology

Figure 1 displays the REST-related methodological option which:
- positions the choice between rules and ML systems prior to any IE algorithm development;
- considers this choice in a differentiated way at the level of each entity;
- considers rules as a default option;
- is based on a short-term analysis of the entities in the text corpus and avoids extended expert annotation.

This method for choosing the optimal IE option relies on the following steps:
1) To highlight by a field expert in a single working session a representative subset of the development set for all types of entities. An entity highlightment relates to the shortest continuous text sample necessary to make the entity be defined and described exhaustively, independently of the text context. An entity highlightment should make sense to another field expert by itself, with no reference to the text corpus. An entity highlightment is quicker to perform than an entity annotation as it does not consider the text disruptions and as it mimics the modalities of human oral communication. For example, the 'tumor size' entity would be highlighted by "the size of the tumor, which shows a low level of differentiation and multiple tumor perineural invasion, is 34 mm", while this entity would be annotated as follows: "34 mm" which does not make sense on its own.

2) To analyze such text highlights in terms of:
- quantitative distribution within the data corpus. The higher the lexical diversity is, the bigger the terminology is needed to model an entity, and thus the less a rule-based method would be considered. As rules are based on pre-existing or homemade terminologies, which limit the number of entries, this method requires that the entity should be formalized within a short list of word expressions.
- qualitative categorization defined by a lowest common semantic denominator. Such step allows the expert to visualize the contribution of each highlight category in terms of expected rule-related recall and precision per entity simultaneously. To approximate the expected rule-related recall, the content of the text highlights should be considered relatively homogeneous, i.e., belonging to limited, well-defined lexical fields and similar syntactic structures. For example, the presence of unlimited synonyms or surrogate terms would be a red flag for rules (e.g., the name of disease signs and symptoms). To approximate the expected rule-related precision, the text highlights should be specific to the entity, i.e., not spread elsewhere in the text (e.g., the mention of generic terms such as "treatment" in the medical field).

3) To conclude by determining whether rule-based IE is a reliable and feasible, at the level of each entity.

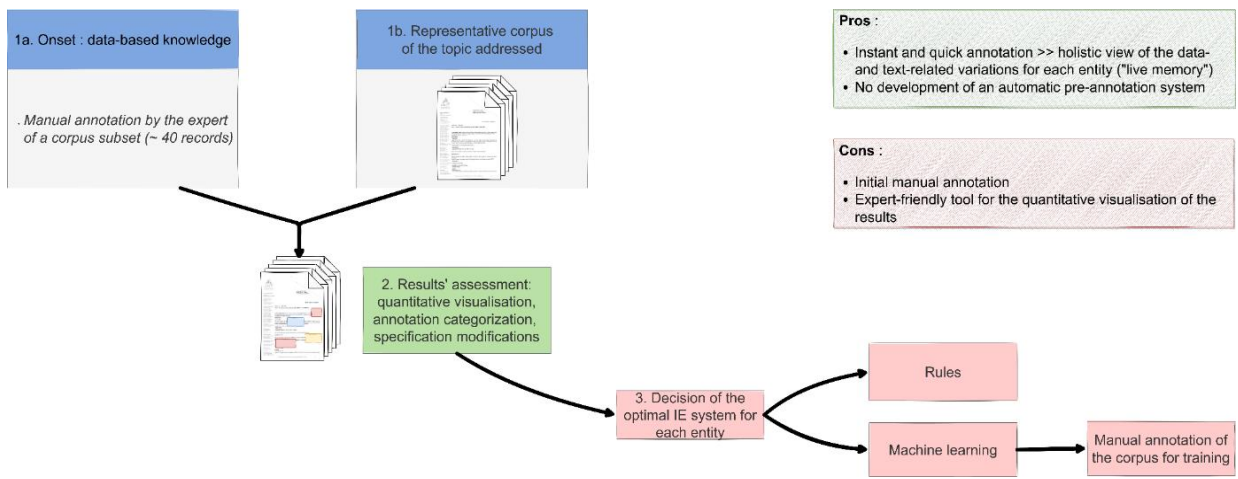

Figure 1: Proposal of an information extraction method combining both rules and machine learning systems, based on the initial assessment of an expert highlighted representative text corpus subset.

## 3. REST interface development

### 3.1 Specifications of the tool

To consider using rule-based method for IE, it is necessary to study the terminology of each entity and consider whereas it could be formalized within a defined, specific and limited list of word expressions. If those word expressions are both precise and specific enough, then rules may be considered for an IE task. Figure 2 and the YouTube video tutorial (https://youtu.be/v58IqJxVnCo) summarize the specifications of REST.

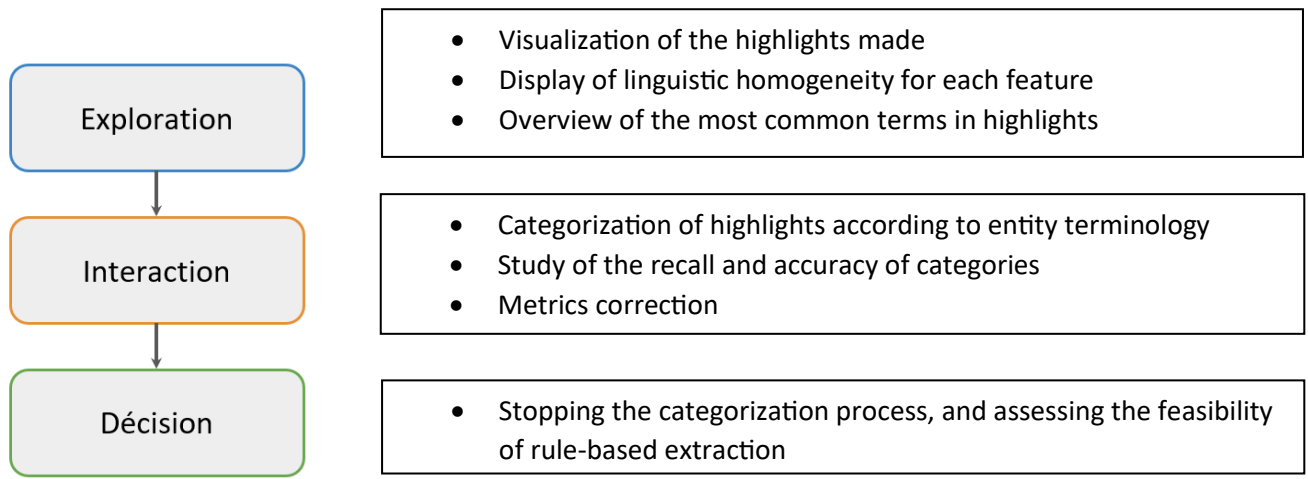

Figure 2: REST specifications and steps of use

- Exploration step

REST allows the user to visualize the entity highlights realized during the initial highlightment task performed by the field expert. REST calculates and presents a linguistic homogeneity score of the text highlights for each entity. The tool calculates and displays the most frequent N-grams included in the text highlights for each entity, as well as a concordancer.

- Interaction step

From the expressions gathered during the exploration phase, the user is supposed to categorize most of the text highlights of each entity, according to their semantic or lexical similarity, mimicking a terminology, with the smallest number of distinct categories. Indeed, for rules, the terminology should be exhaustive and limited to ensure good recall metrics. Then, a REST module enables the user to check that the text highlight categorization does not induce too many false positives, to ensure good precision metrics. In this step, the user can identify the potential text highlight errors.

- Decision step

To decide whether rule-based method for an entity can be considered, a dedicated checklist was developed, based on the prior qualitative and quantitative criteria. We provide the example of a 12-entity IE use case (see below). In the following paragraphs, we will illustrate the REST specifications and steps of use thanks to a 12-entity use case of IE.

### 3.2 Exploration step
#### 3.2.1 Visualization of each entity-related text highlight linguistic homogeneity

To assess the linguistic homogeneity of the text highlights, we implemented a homogeneity score calculation (Figure 3). The goal of such a score is to provide the user with a glimpse of the complexity to build rules for each entity, and therefore to avoid her to start such a time-consuming task. For each entity, a score ranging from 0 to 1 is calculated, representing the lexical diversity of words present in the related highlights. To calculate the score of an entity, all occurrences of each word presented in the corresponding highlights are computed. Then, REST calculates a ratio between the total number of words and the total number of unique words. The homogeneity score $H_e$ for each entity $e$ is calculated as follows:

$$H_e = \frac{T_e - U_e}{T_e},$$

where $T_e$ is the total number of words for an entity $e$ and $U_e$ is the number of unique words for the entity $e$. Finally, to enhance the interpretability of the user, a sigmoid transformation $\sigma$ is performed to increase the discrepancy between the obtained scores, as follows:

$$\sigma(x, k) = \frac{1}{1+e^{-kx}},$$

where $k$ (=10) is a parameter that controls the steepness of the sigmoid curve, resulting on the following transformed homogeneity score: $H_e^\sigma = \sigma(H_e, k)$.

| entity | homogeneity |
|---|---|
| histologie_tumorale | 0.79 |
| traitement_specifique_du_cancer | 0.79 |
| signes_physiques | 0.29 |
| evolutivite_en_lien_avec_le_cancer | 0.01 |
| reponse_a_la_chimiotherapie | 0.82 |
| stade_metastatique_avec_localisations | 0.6 |
| statut_tabagique | 0.62 |
| atcd_geriatriques_et_medicaux_significatifs_pour_la_prise_en_charge | 0.11 |
| stade_oms_ecog_karnofsky | 0.91 |
| biomarqueurs_therapeutiques | 0.67 |
| topographie_du_primitif | 0.57 |
| symptomes | 0.4 |

Figure 3: Visualization of the linguistic homogeneity score results of the text highlights related to the use case entities.

### 3.2.2 Visualization of the most frequent text-highlight N-grams per entity

This REST module is provided to reduce the time spent on the text highlight categorization for each entity. To help the user to identify the most frequent terms characterizing each entity, the tool provides a selection of recommended terms and expressions from the related text highlights (Figure 4). The tool presents the top 10 words from the highlights with the highest term frequency-inverse document frequency (tfidf) score, where:

$$TF.IDF_{t,e} = TF_{t,e} \times IDF_t.$$

In our application of the calculation method, the scale of the text of interest is not the document but the text highlights of each entity. The resulting modified tfidf score is then defined from the two following calculations:

$$mTF_{t,e} = \frac{f_{t,e}}{\sum_{t' \in e} f_{t',e}}, mIDF_t = log\left(\frac{|E|}{|E_t|}\right).$$

For the modified term frequency $m(TF_{t,e})$, the modified $f_{t,e}$ is the occurrence of the term $t$ in the text highlights of the entity $e$, and the denominator represents the total number of occurrences of all the terms in the text highlights of the entity $e$. Regarding the modified inverse document frequency $m(IDF_t)$, $|E|$ is the total number of entities in the corpus and $|E_t|$ is the number of entities containing the term $t$.

Using this technique guarantees that the resulting words are not only frequent, but also more specific of the entity text highlights. If the user estimates that one of them is not accurate or useful, she is able to discard it, allowing the tool to return the next term in the score ranking. Moreover, an entity may be based on N-grams (and not solely on unique words). That's why, for each top 10 words in the tf-idf ranking, the tool is calculating every possible and existing N-grams, ordering the results by their frequency, and returning the top 5 N-grams.

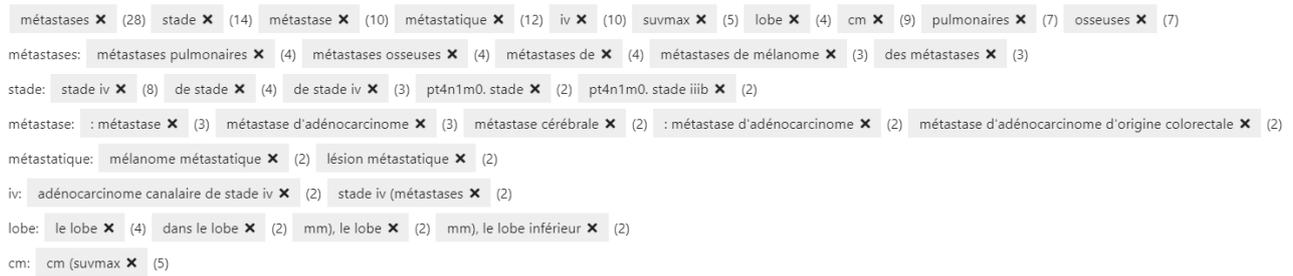

Figure 4: Recommended terms generated for the entity 'metastatic localizations'

#### 3.2.3 Concordancer

We implemented a concordancer, helping the user to identify words or expressions within the whole text corpus (Figure 5). For each resulting match between the typed word and the text, the concordancer indicates if it belongs to an entity highlight (and which one) or not.

Figure 5: Visualization of the concordancer for the word 'tumor'

### 3.3 Interaction step
#### 3.3.1 Text highlight categorization

The user is supposed to categorize most of the text highlights of each entity, according to their semantic similarity, mimicking terminology, with the smallest number of distinct categories. For every category created (corresponding to one or multiple terms), REST creates regular expressions which are supposed to identify all the text highlights belonging to this category. To create highlight categories, the user enters the list of terms specific to each text highlight category, by selecting and sliding the displayed N-grams (Figures 4 and 6). The user may also create regular expressions to build her categories.

Figure 6: Example of 7 categories created by a user for the entity 'tumor histology'.

The uncategorized text highlights are displayed in another REST module (Figure 7). The analysis of the linguistic and semantic of the uncategorized text highlights may help the user to define new categories, or to anticipate that rules are not a feasible IE option due to high linguistic and semantic heterogeneity.

Figure 7: Visualization of the uncategorized text highlights for the entity 'anticancer treatment'

- **Two available features for the text highlight categorization**
- **Spacing word and optimal spacing**

To enhance the flexibility of creating categories, the user can refer to discontinued term expressions. For example, the text highlight 'large cell carcinoma' may contain two discontinued term expressions: the first one refers to "large cell" and the second one refers to "carcinoma". REST enables the user to include a fluctuating number of terms separating both term expressions of interest: the user may type 'large cell …X carcinoma', where X represents the maximum number of characters allowed between the two term expressions of interest (Figure 8A).

Figure 8A: Example of discontinued expressions with a maximum distance of 15 between two terms.

The tool calculates all the matches possible of such a discontinued term expression in the whole text corpus, regardless of the distance between the term expressions of interest. If this match corresponds to an existing highlight, it will be considered as a true positive (TP) or as a false positive (FP), according to the expert text highlightment (Figure 8B). Therefore, the user can adjust the distance entered based on the balance of TP and FP that she will catch for the word expression.

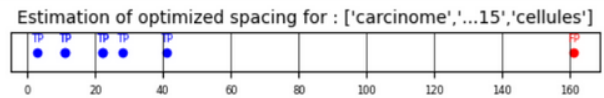

Figure 8B: Calculation of optimal spacing for the word expression 'carcinoma …15 cells'

- **Ban words identification**

Another REST module allows the user to decrease the number of FP cases, due to polysemic words. For example, medical reports make a distinction between the patient's and the family's medical background. Thus, typing 'background' in a patient-related category would gather unwanted FP regarding the family's medical background. By using the ban words expression "family background", REST may discard the FP containing this expression, resulting in an increase of the category precision. Moreover, banned expressions are easily deployed in regular expressions, making their use here meaningful.

### 3.3.2 Visualization of the precision of each text highlight category

REST enables the assessment of the precision of each highlight category by the following method. After categorizing the highlights, the same underlying regular expressions are applied on the whole text corpus. If the match corresponds to an existing highlight, it will be considered as TP. The precision for the category can then be computed (Fig. 9A), as well as the overall precision of the entity (i.e., all the categories taken together) (Fig. 9B).

**II - Categories metrics results**

| category | raw highlights | | | | corrected highlights | | | |
|---|---|---|---|---|---|---|---|---|
| | TP | FP | FN | precision | TP(corr) | FP(corr) | FN(corr) | precision |
| 'métastase', 'métastatique' | 22 | 11 | 0 | 0.67 | 26 | 7 | 0 | 0.79 |
| 'stade iv', 'stade 4', 'stade IV' | 9 | 2 | 0 | 0.82 | 9 | 2 | 0 | 0.82 |
| 'm0', 'm1' | 3 | 1 | 0 | 0.75 | 4 | 0 | 0 | 1 |
| 'implants péritonéaux', 'implant péritonéal' | 3 | 0 | 0 | 1 | 3 | 0 | 0 | 1 |
| 'carcinose' | 1 | 8 | 0 | 0.11 | 6 | 3 | 0 | 0.67 |
| 't.n.m.', 'pt.n.m.', 'ct.n.m.', 'pt..n.m.' | 7 | 4 | 0 | 0.64 | 9 | 2 | 0 | 0.82 |

Figure 9A: Performance metrics of the 6 created categories related to the text highlights of the 'metastatic stage with localization' entity.

**I - Entity metrics results**

| key | entity | homogeneity | TP | FP | FN | precision | precision_confide | recall | recall_confidence |
|---|---|---|---|---|---|---|---|---|---|
| 5 | stade_metastatique_avec_localisations | 0.6 | 57 | 14 | 38 | 0.8 | [0.69, 0.9] | 0.6 | [0.49, 0.71] |

Figure 9B: Performance metrics of the created categories at the scale of the 'metastatic stage with localization' entity.

### 3.3.3 Updating text highlight-related performance metrics

Mistakes in the initial text highlight expert task may occur and bias the metrics defined previously and should be considered. To allow the user to have a retroactive effect on the text data, REST provides a dedicated feature (Figure 10). She can visualize the term expressions from the whole text corpus which match with the categories created and she can check whether the related text sample was well classified (TP vs. FP). FP means that text highlights were missed by the expert. The tool user may convert each misclassified text highlight manually, which results in automatically updated metrics.

...clusivement pleurale et ganglionnaire, se prêtant à un traitement par chimiothérapie (ct) à visée palliative. traitement il a commencé un traitement e...

☐ Consider this FP as a TP

| key | category | result | text | file | places |
|---|---|---|---|---|---|
| 0 | 'chimiothérapie' | TP | ...ec une étude d'extension négative. le patient reçoit un traitement de chimiothérapie néoadjuvante avec un schéma ... | cc_onco1000... | 2229,2243 |
| 1 | 'chimiothérapie' | TP(corr) | ...xcellente réponse pathologique du carcinome canalaire infiltrant à la chimiothérapie, une biopsie de la paroi thoraci... | cc_onco1000... | 3270,3284 |
| 2 | 'chimiothérapie' | TP(corr) | ... d'environ 10 cm au niveau de la cuisse droite, une nouvelle ligne de chimiothérapie commence donc avec r-gemox... | cc_onco1000... | 3746,3760 |
| 3 | 'chimiothérapie' | TP(corr) | ...et le suivi. compte tenu du diagnostic susmentionné, un traitement de chimiothérapie a été débuté avec le schéma ... | cc_onco102.txt | 2131,2145 |
| 4 | 'chimiothérapie' | FP | ...lement avancé ou métastatique après échec d'au moins un traitement de chimiothérapie antérieur. cela commence ... | cc_onco102.txt | 3505,3519 |
| 5 | 'chimiothérapie' | FP | ...clusivement pleurale et ganglionnaire, se prêtant à un traitement par chimiothérapie (ct) à visée palliative.   traitem... | cc_onco103.txt | 2346,2360 |
| 6 | 'chimiothérapie' | TP | ...re médiastinale. depuis un an et demi, il bénéficie de traitements de chimiothérapie à base de platine, en raison de l... | cc_onco103.txt | 4887,4901 |

Figure 10: Example of misclassified text highlight modification.

*The user selects a line in the table and can then decide to consider this FP as a TP if it corresponds to a misclassified item by checking the above box.*

Looking at the right part of Figure 9A, the highlight correction step increased the precision for some of the categories of the entity 'metastatic stage with localization'. It results in an increase of the entity precision from 0.63 to 0.8 (Fig. 9B).

## 3.4 Decision step

### 3.4.1 When to stop the text highlighting categorization?

The user should optimize the number of categories (i.e., to create the minimal number of distinct categories related to the most text highlights). The main goal behind rule-based IE method is to capture the biggest amount of information of interest with the smallest number of regular expressions. In other terms, the highest recall is to be obtained from the smallest number of categories.

Figure 11 displays the text highlighting category recall results for two distinct entities. For the 'smoking status' entity, 3 categories could capture 100% of the highlights, and each of these categories captured from 14 to 64% of the entity highlights (Figures 11a and 11c). Conversely, for the 'physical signs' entity, each of the 4 created categories made of terms with the highest modified tf-idf score captured 6.85% of the text highlights (Fig 11b and 11d). The following categories would have even less highlight "recall" than the previous 4, because of a decreasing modified tf-idf. This would result in a long list of categories capturing a smaller portion of the text highlights, which should be avoided for rule creation.

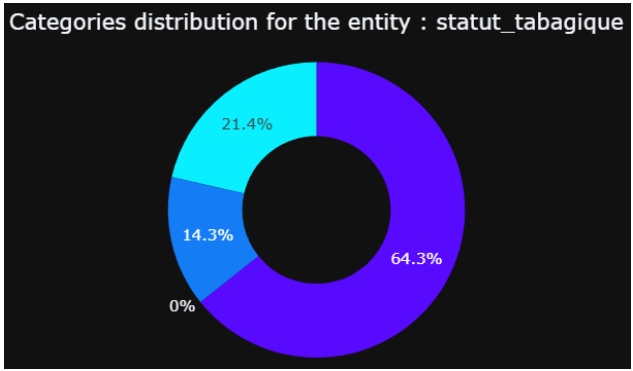
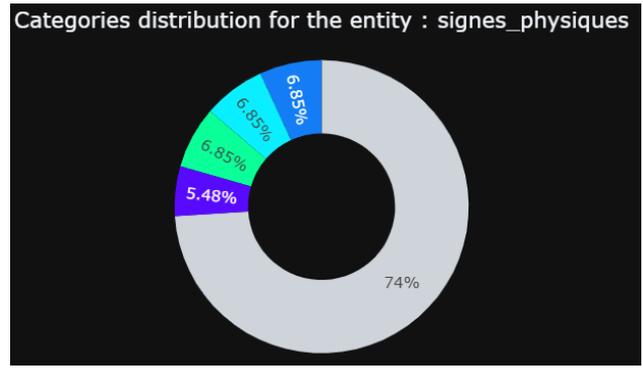

(a) Recall results for the entity 'smoking status'

(b) Recall results for the entity 'physical signs'

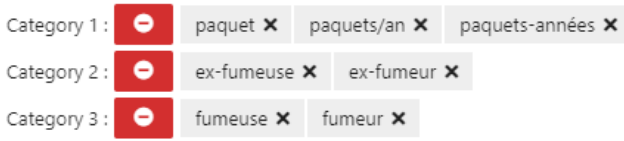
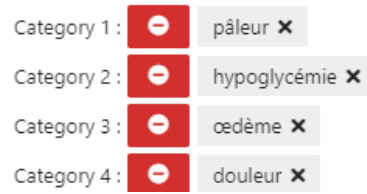

(c) Categories of the entity 'smoking status'

(d) Categories of the entity 'physical signs'

Figure 11: Two examples of category creation recall on two different entities, and their related categories.

*The grey part (Fig. 11a and 11c) represents the uncategorized text highlights*

### 3.4.2 When not to consider rules as an IE method option?

As a final step, REST provides the user with a checklist summarizing the quantitative and qualitative criteria assessed and visualized with such a tool (Figure 12).

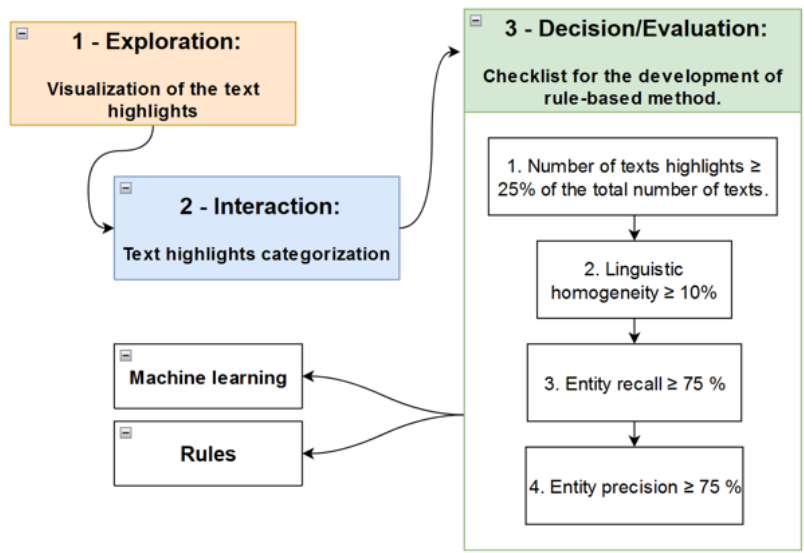

Figure 12: Exploration, interaction and decision steps as offered by REST, including the checklist criteria to help the user to choose between rules and ML as an optimal IE method.

## 4. REST evaluation
### 4.1 Corpus characteristics, preparation and gold standard

We used an open access datataset named CANTEMIST, containing a corpus of 500 Spanish electronic health care records in oncology. We then translated those texts into French by using the Google translate package available on jupyter. We selected 12 entities belonging to the cancer minimal dataset, and a junior field expert (FA) highlighted those entities on 35 reports of the corpus (15). Then, as a gold standard, a senior field expert (EK) went through the categorization process for all the entities in a total of 142 min, and she gave a final opinion about the feasibility of rule development per entity

### 4.2 External evaluations

The assessment of rule feasibility was realized by the same junior field expert (FA) that had no annotation knowledge nor NLP previous experience. She took 130 minutes to perform an evaluation on all the entities, using the same checklist of rule feasibility.

Table 1: Summary of rule feasibility as assessed by the senior (EK) and the junior field expert (FA), for the 12 entities.

| Senior Evaluation \ Junior Evaluation | Validated Rule application | Ø Validated Rule application | Total |
|---|---|---|---|
| Validated Rule application | **5** | **0** | 5 |
| Ø Validated Rule application | **1** | **6** | 7 |
| Total | 6 | 6 | 12 |

Among the 12 evaluated entities, the junior and senior expert agreed on 11 entities, resulting in an overall agreement of 91.67% (Table 1). A divergence was noted for the entity 'Therapeutic Biomarkers' (Table 2). This divergence lied in the categorization method understanding: since the desired information to extract was the value of the biomarkers, the senior expert decided that putting the biomarkers alone in the categories were not sufficient for the creation of future regular expressions. This example highlights the importance of a predefined highlight guide.

## 5. Conclusion

To improve the sustainability of IE algorithm development and validation, we hypothesized that rules could be a reliable IE method option, depending on text characteristics at the level of each entity and that rules could be combined to ML in this setting. We developed and validated the user-friendly visualization and decision aid REST to help NLP agnostic annotators to assess the feasibility of rules for a dedicated IE task, providing them with a tutorial video and an external validation.

Table 2: Decision for rules for each entity of the use case, as assessed by a senior (EK) and a junior (FA) field expert

| Entity | Criteria | Senior | Junior |
|---|---|---|---|
| Tumor Histology | TH ≥ 25% | Yes | Yes |
| | LH ≥ 10% | Yes | Yes |
| | ER ≥ 75% | Yes | Yes |
| | EP ≥ 75% | Yes | Yes |
| Specific Cancer Treatment | TH ≥ 25% | Yes | Yes |
| | LH ≥ 10% | Yes | Yes |
| | ER ≥ 75% | Yes | Yes |
| | EP ≥ 75% | Yes | Yes |
| Physical Signs | TH ≥ 25% | Yes | Yes |
| | LH ≥ 10% | Yes | Yes |
| | ER ≥ 75% | No | No |
| | EP ≥ 75% | - | - |
| Cancer-Related Progression | TH ≥ 25% | No | No |
| | LH ≥ 10% | - | - |
| | ER ≥ 75% | - | - |
| | EP ≥ 75% | - | - |
| Chemotherapy Response | TH ≥ 25% | Yes | Yes |
| | LH ≥ 10% | Yes | Yes |
| | ER ≥ 75% | Yes | Yes |
| | EP ≥ 75% | Yes | Yes |
| Metastatic Stage and Localizations | TH ≥ 25% | Yes | Yes |
| | LH ≥ 10% | Yes | Yes |
| | ER ≥ 75% | No | No |
| | EP ≥ 75% | - | - |
| Smoking Status | TH ≥ 25% | Yes | Yes |
| | LH ≥ 10% | Yes | Yes |
| | ER ≥ 75% | Yes | Yes |
| | EP ≥ 75% | Yes | Yes |
| Significant Geriatric and Medical History | TH ≥ 25% | Yes | Yes |
| | LH ≥ 10% | No | No |
| | ER ≥ 75% | - | - |
| | EP ≥ 75% | - | - |
| WHO ECOG Karnofsky Score | TH ≥ 25% | Yes | Yes |
| | LH ≥ 10% | Yes | Yes |
| | ER ≥ 75% | Yes | Yes |
| | EP ≥ 75% | Yes | Yes |
| **Therapeutic Biomarkers** | TH ≥ 25% | Yes | Yes |
| | LH ≥ 10% | Yes | Yes |
| | **ER ≥ 75%** | **No** | **Yes** |
| | EP ≥ 75% | Yes | Yes |
| Primary Tumor Topography | TH ≥ 25% | Yes | Yes |
| | LH ≥ 10% | Yes | Yes |
| | ER ≥ 75% | No | No |
| | EP ≥ 75% | - | - |
| Symptoms | TH ≥ 25% | Yes | Yes |
| | LH ≥ 10% | Yes | Yes |
| | ER ≥ 75% | No | No |
| | EP ≥ 75% | - | - |

Abbreviations: **ER**: Entity Recall, **EP**: Entity Precision, **LH** Linguistic homogeneity. **TH**: number of Text Highlights